\documentclass{article} % For LaTeX2e
\usepackage{iclr2026_conference,times}

\usepackage{amsmath,amsfonts,bm}

\def\eqref#1{equation~\ref{#1}}
\def\1{\bm{1}}

\DeclareMathAlphabet{\mathsfit}{\encodingdefault}{\sfdefault}{m}{sl}
\SetMathAlphabet{\mathsfit}{bold}{\encodingdefault}{\sfdefault}{bx}{n}

\usepackage[utf8]{inputenc} % allow utf-8 input
\usepackage[T1]{fontenc}    % use 8-bit T1 fonts
\usepackage{hyperref}       % hyperlinks
\usepackage{url}            % simple URL typesetting
\usepackage{booktabs}       % professional-quality tables
\usepackage{amsfonts}       % blackboard math symbols
\usepackage{nicefrac}       % compact symbols for 1/2, etc.
\usepackage{microtype}      % microtypography
\usepackage{xcolor}         % colors
\usepackage{graphicx}
\usepackage{makecell}
\usepackage{subcaption}

\title{Reusing Pre-Training Data at Test Time is a Compute Multiplier}
\author{%
  \quad \textbf{Alex Fang}$^{2,3}$
  \quad \textbf{Thomas Voice}$^{1}$
  \quad \textbf{Ruoming Pang}$^{2}$
  \quad \textbf{Ludwig Schmidt}$^{3}$
  \quad \textbf{Tom Gunter}$^{2}$ \\
\quad $^1$Apple \quad $^2$ work done while at Apple\quad $^3$Stanford
}

\iclrfinalcopy % Uncomment for camera-ready version, but NOT for submission.
\begin{document}

\maketitle

\begin{abstract}
% Large language models learn from their vast pre-training corpora to gain the ability to solve an ever increasing variety of tasks; yet, pre-training data is still underutilized. In this work, we show that retrieving from standard pre-training datasets at test time can lead to large gains on both knowledge and reasoning tasks. We explore combining retrieval with voting through inter-document consistency, resulting in an 8.7 percentage point gain on MMLU and a 17.9 percentage point gain on MATH-500. We further explore the relationship between pre-training and retrieval, finding that (TODO).
Large language models learn from their vast pre-training corpora, gaining the ability to solve an ever increasing variety of tasks; yet although researchers work to improve these datasets, there is little effort to understand how efficient the pre-training apparatus is at extracting ideas and knowledge from the data. In this work, we use retrieval augmented generation along with test-time compute as a way to quantify how much dataset value was left behind by the process of pre-training, and how this changes across scale.
We demonstrate that pre-training then retrieving from standard and largely open-sourced datasets results in significant accuracy gains in MMLU, Math-500, and SimpleQA, which persist through decontamination. For MMLU we observe that retrieval acts as a $\sim5$x compute multiplier versus pre-training alone. 
We show that these results can be further improved by leveraging additional compute at test time to parse the retrieved context, demonstrating a 10 percentage point improvement on MMLU for the public LLaMA 3.1 8B model.
% Our results suggest that with current methods, pre-training on existing pre-training datasets is not making full use of the information they contain, leaving room for substantial improvements in many areas of pre-training research.
Overall, our results suggest that today's pre-training methods do not make full use of the information in existing pre-training datasets, leaving significant room for progress.
\end{abstract}

\section{Introduction}
% Old, out of date

Large language models (LLMs) have consistently improved performance by scaling pre-training compute \citep{hestness2017deep,hoffmann2022training,kaplan2020scaling}. In parallel to scaling, researchers have also made significant efforts to improve both the architectures and the datasets used by these LLMs \citep{gu2023mamba, li2024datacomp,raffel2020exploring,shazeer2017outrageously}. While LLMs are able to solve an incredible number of tasks, in their current form they have several limitations. For example, they struggle with long-tail knowledge \citep{kandpal2023large} and have limitations in their ability to generalize, such as for the reversal curse \citep{berglund2023reversal}. Additionally, they have been observed to have a log-linear scaling trend, meaning that larger amounts of compute are necessary to make the same gains at larger scales. To understand whether these limitations come from the quality of the datasets, it is important to explore whether further improvements can be unlocked by using them beyond pre-training.

% There has been significant effort to extend LLM capabilities by increasing and improving pre-training datasets; however, it is not clear that these limitations are caused by insufficient knowledge in the training data.

Taking advantage of the effort put into creating these sophisticated pre-training datasets, we explore whether reusing them through retrieval at test time can further improve performance. Additionally, we test if supplementing with additional test-time compute results in further gains. Extra test-time compute can be applied naturally to retrieval augmented generation by running multiple trials for self-consistency while changing the retrieved documents across different trials. We apply this variety of techniques on a set of publicly available pre-training datasets, to measure the potential impact of the knowledge contained in them.

First, we pre-train models at various compute budgets and then use the same dataset for retrieval. We see that retrieval further benefits our models when evaluating on MMLU, Math-500, and SimpleQA, even though the models were pre-trained on the same data.
We fit a power law to the base pre-training models across compute budgets and compare against the gain from retrieval, showing that while on average retrieval provides a $\sim5$x compute multiplier over pre-training when evaluating on MMLU, the effectiveness degrades with scale.

Next, we use additional test-time compute on the pre-training data with a combination of retrieval and self-consistency techniques. We evaluate our methods on a variety of downstream tasks that cover multiple domains, requiring both knowledge and reasoning abilities. With a Llama 3.1 8B reader model, we use retrieval to achieve 74.0\% on SimpleQA, as well as a 10.5 percentage point gain on MMLU, a 15.7 percentage point gain on MATH-500, and a 6.2 percentage point gain on GPQA.

Lastly, we analyze our findings to provide suggestions for further improving these datasets. We are able to relate performance gaps between controlled experiments back to specific stages of the dataset creation pipeline. Altogether, our work suggests that there is room for improving both pre-training datasets and learning methods using these datasets.
% TODO (add conclusion given relationship between pre-training and retrieval)
% more about high level impact or implications? e.g. following paragraph from test time section
% If we view retrieval as a tool for the LLM, then our methods use test-time compute to improve the tool itself. This contrasts with self-consistency by itself, which parallelizes the model without additional enhancement, as well as with deep research, which uses test-time compute to use the tool for longer rather than to upgrade it. Retrieval is our vehicle through which we can put in additional compute, and do it in a data driven way.

\section{Related Work}
% Old, maybe out of date
% other ideas for things to add? maybe pre-training dataset construction or diferent ways to use data?
Classical pre-training scaling law studies established relationships for how loss falls with training compute, data, and parameters \citep{hestness2017deep, kaplan2020scaling, hoffmann2022training}. More recent analyses factor in deployment costs in high-inference-demand settings \cite{beyondchinchilla}.

Meanwhile, retrieval-augmented approaches externalize knowledge to non-parametric memory \citep{guu2020realmretrievalaugmentedlanguagemodel, lewis2020retrieval, ram2023context}, trading additional test time compute for improved performance on knowledge-intensive tasks. More recently, \citet{shao2024scaling} demonstrated that scaling up the datastore can reliably further improve performance on knowledge-intensive tasks, while \citet{lyu2025frustratingly} demonstrated that a compact subset of pre-training data can be used in a practical way with a minimal retrieval setup to improve performance on reasoning benchmarks. 

Recent work \citep{brown2024large,snell2024scaling} has shown that scaling test-time compute can be an efficient way of improving LLM performance. Specifically, they suggest parallelizing inference compute across trials, and sequentially iterating on a model's output. Results can then be aggregated with techniques like self-consistency \citep{wang2022self, chen2023universal} or verifiers. % is there a good cite for verifiers?
Retrieval is naturally suited to both as it can be parallelized across retrieved documents, while sequentially iterating over the search query to improve the retrieved documents.

Large-scale commercial "Deep Research" systems from companies like Google, OpenAI, and Perplexity likely apply all of the above by abstracting retrieval behind tool-use APIs and allowing the model direct control over the tools along with additional test-time compute.

In this paper, we first take inspiration from the classical pre-training scaling law studies, and begin to characterize the joint scaling of pre-training and simple retrieval given a fixed and identical data corpus. We also explore which forms of simple test-time compute allow us to most effectively leverage the retrieval stores, and whether these conclusions generalize to settings where the pre-training corpus differs from the retrieval datastore.

\section{Experimental Setup}
\label{sec:setup}

\subsection{Datasets}
\begin{table}[ht]
\centering
\caption{Pre-training datasets used. Tokens refers to the actual size of the dataset, all of which is typically used during retrieval. Epochs refers to the number of epochs taken for the largest pre-training run. Epochs for smaller models are scaled down proportionally.}
\begin{tabular}{lrr}
\toprule
\textbf{Dataset} & \textbf{Tokens} & \textbf{Epochs} \\
\midrule
\multicolumn{3}{l}{Web Crawl Based} \\
\midrule
DCLM-baseline (dedup) \citep{li2024datacomp} & 764.9B & 0.88 \\
FineWeb-edu (dedup) \citep{penedo2024fineweb} & 197.6B & 0.80 \\
\midrule
\multicolumn{3}{l}{Additional Sources} \\
\midrule
arXiv & 28.7B & 1.10 \\
peS2o \citep{peS2o} & 71.9B & 1.31 \\
PubMed Central & 22.5B & 2.72 \\
Stack Exchange & 11.8B & 7.98 \\
Wikipedia & 21.7B & 3.63 \\
\midrule
\multicolumn{3}{l}{Math} \\
\midrule
AlgebraicStack \citep{azerbayev2023llemma} & 9.9B & 6.38 \\
AutoMathText \citep{zhang2025autonomous} & 6.0B & 13.11 \\
FineMath-3+ \citep{allal2025smollm2smolgoesbig} & 36.5B & 1.72 \\
FineMath-4+ \citep{allal2025smollm2smolgoesbig} & 10.0B & 11.05 \\
OpenWebMath \citep{paster2023openwebmath} & 13.5B & 2.91 \\
StackMathQA \citep{stackmathqa2024} & 0.7B & 558.67 \\
\bottomrule
\end{tabular}
\label{tab:datasets}
\end{table}
We use the exact same datasets for both pre-training and retrieval. We include the standard webcrawl based large-scale datasets DCLM-baseline and FineWeb-edu (deduplicated versions for both), followed by more specialized sources. These include arXiv, peS2o, PubMed Central, Stack Exchange, and Wikipedia. Lastly, we include AlgebraicStack, AutoMathText, FineMath-3+, FineMath-4+, OpenWebMath, and StackMathQA to improve mathematics coverage. Additional information on the datasets such as token count and pre-training mixing ratios can be found in Table~\ref{tab:datasets}.

\subsection{Retrieval Pipeline}
We use Qwen3 Embedding 0.6B and Qwen3 Reranker 0.6B \citep{qwen3embedding} as the embedding and reranker models. We use FAISS FlatIP \citep{johnson2019billion} for indexing to retrieve the top 100 documents per dataset (or shard of a dataset) before combining across all datasets through sorting by similarity score. This is equivalent to having all datasets in the same index and retrieving the top 100, but is more practical due to compute constraints. Then if applicable, we rerank the top 100 documents across all datasets to achieve our final document ordering.

\section{Pre-Training Experiments}
\label{sec:pretrain}
\begin{figure*}[htbp]
    \centering
    \includegraphics[width=1.0\linewidth]{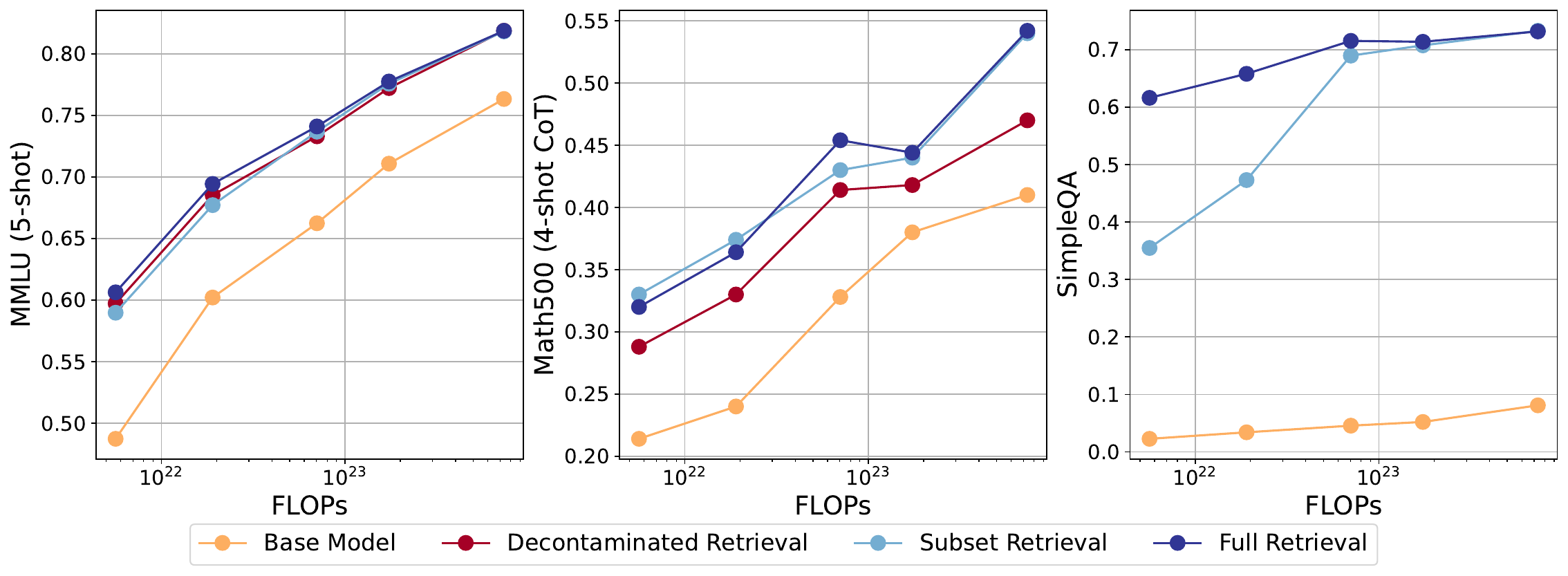}
    \caption{Retrieval on the pre-training dataset can substantially improve upon the performance of the base model. However, the exact benefit depends on the type of task.}
    \label{fig:figure1}
\end{figure*}

We aim to measure the knowledge contained in pre-training datasets by first pre-training on it, then also retrieving (with reranker) on it during test time. In Figure~\ref{fig:figure1} we measure performance on MMLU \citep{hendrycks2020measuring}, Math-500 \citep{lightman2023let}, and SimpleQA \citep{wei2024measuring} across compute budgets, comparing the base model with retrieval on all our datasets, retrieval on a decontaminated version of all our datasets, and retrieval on a subset approximately equivalent to the unrepeated pre-training dataset. Overall, we find that retrieval can help on all three tasks, though to different degrees.
% How should we metnion pre-training details here and point to Appendix~\ref{app:pretraining}?
% Mention three types of evaluation tasks, differences between them (general knowledge, mathematical reasoning, factual knowledge, broadly retrieval always helps

Retrieving from the full dataset leads to large gains on all tasks, but the model sees more data than the base model sees in pre-training. We account for this by retrieving from a subset similar to the unrepeated pre-training dataset, represented by the light blue lines in Figure~\ref{fig:figure1}. Interestingly, this achieves similar performance to retrieving on the full dataset for MMLU and Math-500. On the other hand, performance on SimpleQA is a piecewise function because SimpleQA depends heavily on retrieving from Wikipedia, and the smaller two models only see a fraction of Wikipedia while the larger three models see at least one epoch of Wikipedia. 
% subset retrieval doesn't matter that much except for simpleqa, explain why
% caveat about random subsets, mention proper experiment for 6.4B (similar numbers to decontaminated) Appendix~\ref{app:detokenize_subset}
% any other interesting comments or analysis on subset?

% FLOPs multiplier for MMLU because standard pre-training evaluation
\paragraph{Retrieval as a compute multiplier}
Given the common use of MMLU as a proxy for pre-training quality, we use it to measure the effect of retrieval as a compute multiplier for the base model. We fit a bounded sigmoid function to the base model's MMLU performance as a function of FLOPs, and then measure the amount of pre-training compute needed to match each existing base model augmented with retrieval. In Table~\ref{tab:multiplier} we find that the average compute multiplier across the five models of different scales is 4.86x, and the geometric mean is 4.66x. However, this compute ratio decreases as the model scales, with retrieval providing only a 2.88x compute multiplier at the largest scale. While retrieval provides large compute savings, as the base model is scaled up, retrieval faces greater diminishing returns than just the base model. It is important to notice that retrieval does not provide a flat or strictly decreasing benefit across compute budgets. There is an initial increase in retrieval efficiency, suggesting that it benefits from better base models.

\iffalse
\begin{table}[h]
\centering
\caption{We fit the base model performance to a power law with offset to get $y = 8.25 \times 10^{-6} \times x^{0.1046} + 0.1573$. We use this equation to measure retrieval as a compute multiplier for the base model. The average compute ratio is 6.19x, the geometric mean is 5.50x, and the median is 6.37x.}
\begin{tabular}{ccccc}
\toprule
Compute Budget & \makecell{Baseline \\ MMLU} & \makecell{Retrieval \\ MMLU} & \makecell{Compute for base \\ to match retrieval}  & Compute Ratio \\
\midrule
$5.64 \times 10^{21}$ & 0.4873 & 0.6063 & $3.93 \times 10^{22}$ & 6.97x \\
$1.90 \times 10^{22}$ & 0.6021 & 0.6943 & $1.82 \times 10^{23}$ & 9.56x \\
$7.04 \times 10^{22}$ & 0.6623 & 0.7410 & $4.48 \times 10^{23}$ & 6.37x \\
$1.74 \times 10^{23}$ & 0.7107 & 0.7775 & $8.76 \times 10^{23}$ & 5.05x \\
$7.34 \times 10^{23}$ & 0.7633 & 0.8186 & $2.21 \times 10^{24}$ & 3.01x \\
\bottomrule
\end{tabular}
\label{tab:multiplier}
\end{table}
\fi
\begin{table}[h]
\centering
\caption{We fit the base model performance to a sigmoid function with bounds to get $y = 0.25 + \dfrac{0.6907}{1 + \exp\!\big(-0.7968 \cdot (\log_{10}(x) - \log_{10}(2.48 \times 10^{22}))\big)}$, where 0.25 is the random baseline and 0.9407 is the maximum achievable accuracy \citep{gema2024we}. We use this equation to measure retrieval as a compute multiplier for the base model. The average compute ratio is 4.86x, the geometric mean is 4.66x, and the median is 4.74x.} % maybe 0.9351 is micro average, check what macro should be, maybe 0.941
% macro average should be 0.9407; by sub is STEM: 95.44, humanities 93.77, social sciences 95.75, other 91.14
\begin{tabular}{ccccc}
\toprule
Compute Budget & \makecell{Baseline \\ MMLU} & \makecell{Retrieval \\ MMLU} & \makecell{Compute for base \\ to match retrieval} & Compute Ratio \\
\midrule
$5.64 \times 10^{21}$ & 0.4873 & 0.6063 & $2.98 \times 10^{22}$ & 5.28x \\
$1.90 \times 10^{22}$ & 0.6021 & 0.6943 & $1.36 \times 10^{23}$ & 7.17x \\
$7.04 \times 10^{22}$ & 0.6623 & 0.7410 & $3.34 \times 10^{23}$ & 4.74x \\
$1.74 \times 10^{23}$ & 0.7107 & 0.7775 & $7.35 \times 10^{23}$ & 4.23x \\
$7.34 \times 10^{23}$ & 0.7633 & 0.8186 & $2.11 \times 10^{24}$ & 2.88x \\
\bottomrule
\end{tabular}
\label{tab:multiplier}
\end{table}

% talk about decontamination
\paragraph{Decontamination}
A common question is whether retrieval gains come from retrieving text containing exact overlap with the test data. We decontaminate the retrieved documents for MMLU and Math-500 through n-gram overlap with the questions, as detailed in Appendix~\ref{app:decontamination}. In Figure 1, the red decontaminated retrieval line is close to the dark blue full retrieval line for MMLU, demonstrating that the gains are not attributable to simple contamination. Although Math-500 shows signs of more significant contamination, retrieving against a decontaminated training set still shows a very meaningful improvement over the baseline.
We do note that our analysis shows that 14.1\% of MMLU and 32.0\% of Math-500 can be found in our commonly used open-source pre-training datasets, highlighting the importance of strictly decontaminated (or held out) evaluation sets for pre-training science.
We choose to omit n-gram overlap decontamination analysis for SimpleQA due to the nature of the evaluation task.

\iffalse
\begin{table}[h]
\centering
\caption{Summary of pre-training vs retrieval compute ratios across all MMLU categories. Values show how many times more compute the base model would need to match retrieval performance.}
\label{tab:summary_ratios}
\begin{tabular}{lccccc}
\toprule
Compute Budget & STEM & Humanities & Social & Other & All \\
\midrule
$5.64 \times 10^{21}$ & 4.80× & 5.35× & 4.60× & 10.18× & 6.97× \\
$1.90 \times 10^{22}$ & 5.94× & 8.01× & 7.49× & 13.74× & 9.56× \\
$7.04 \times 10^{22}$ & 2.84× & 6.37× & 7.79× & 7.40× & 6.37× \\
$1.74 \times 10^{23}$ & 2.38× & 5.55× & 5.99× & 4.95× & 5.05× \\
$7.34 \times 10^{23}$ & 1.35× & 2.96× & 3.00× & 2.24× & 3.01× \\
\midrule
Average & 3.46× & 5.65× & 5.77× & 7.70× & 6.19× \\
Geometric Mean & 3.06× & 5.23× & 5.36× & 6.34× & 5.50× \\
\bottomrule
\end{tabular}
\end{table}
\fi
\begin{table}[h]
\centering
\caption{Summary of pre-training vs retrieval compute ratios across MMLU categories. Values show how many times more compute the base model would need to match retrieval performance. Calculated using category-specific bounded sigmoids (min = 0.25; max: STEM 0.9544, Humanities 0.9377, Social 0.9575, Other 0.9114, All  0.9407).}
\label{tab:summary_ratios}
\begin{tabular}{lccccc}
\toprule
Compute Budget & STEM & Humanities & Social & Other & All \\
\midrule
$5.64 \times 10^{21}$ & 6.82x & 2.69x & 3.42x & 9.80x & 5.28x \\
$1.90 \times 10^{22}$ & 10.23x & 3.33x & 4.42x & 13.78x & 7.17x \\
$7.04 \times 10^{22}$ & 5.23x & 2.57x & 4.07x & 8.77x & 4.74x \\
$1.74 \times 10^{23}$ & 5.01x & 2.44x & 3.43x & 7.53x & 4.23x \\
$7.34 \times 10^{23}$ & 3.52x & 1.55x & 2.22x & 6.48x & 2.88x \\
\midrule
Average & 6.16x & 2.52x & 3.52x & 9.27x & 4.86x \\
Geometric Mean & 5.78x & 2.44x & 3.42x & 8.96x & 4.66x \\
Median & 5.23x & 2.57x & 3.43x & 8.77x & 4.74x \\
\bottomrule
\end{tabular}
\end{table}

% \subsection{Breakdown of Impacts of Retrieval Addition and Model Size Increase}
\subsection{Learning from pre-training vs retrieval} % which title better?
\begin{figure}[ht]
  \centering
  % Uncomment and replace with your figure
  \includegraphics[width=1.0\linewidth]{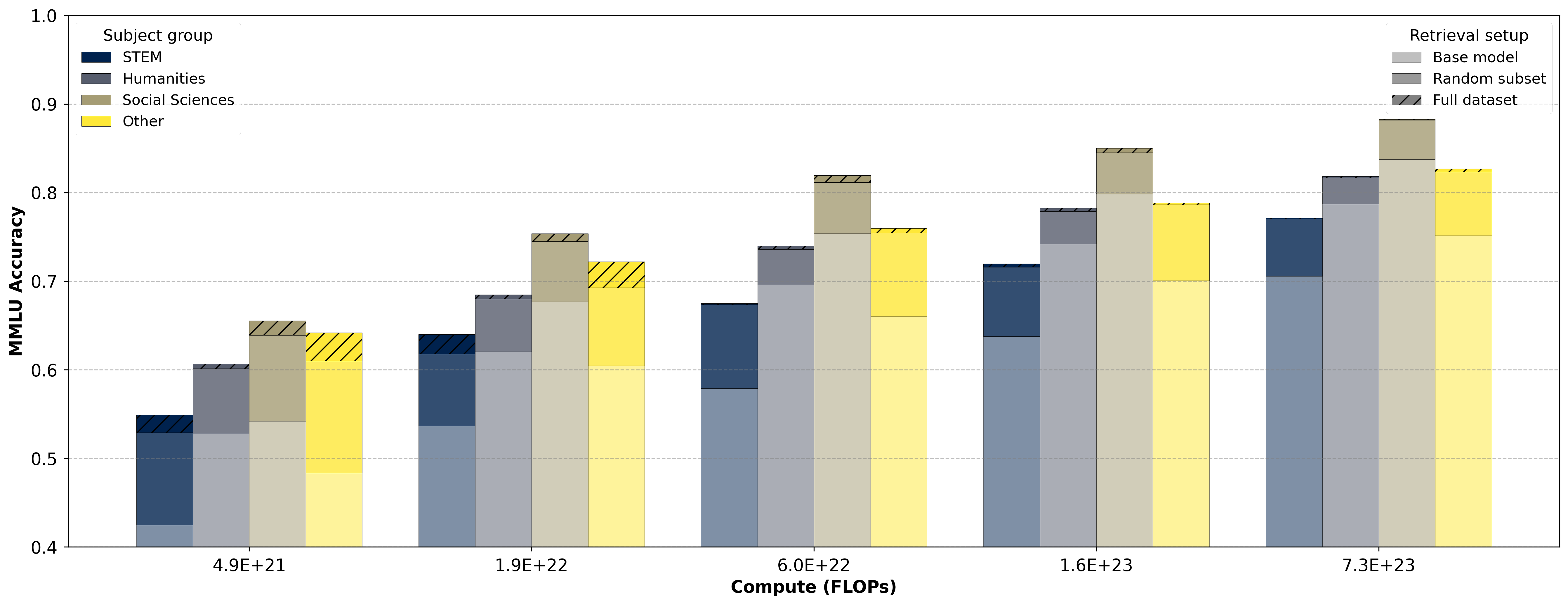}
  %\fbox{\rule{0pt}{4cm}\rule{14cm}{0pt}}
  \caption{MMLU Breakdown by category of impact of retrieval addition and compute budget. Retrieval provides a strong lift, and the difference between retrieving from a random subset of the data store and the full set is small and diminishing with scale.}
  \label{fig:cat_impact}
\end{figure}

In an effort to determine how retrieval can improve model performance, compared to scaling up model size and compute budget, we analyzed the accuracy of the trained models on MMLU, broken down into question categories.
The results in Figure~\ref{fig:cat_impact} show that across categories, the addition of retrieval gives a comparable boost in accuracy to that of a significant increase in pre-training compute budget.

Since retrieval involves a memory storage mechanism, we might expect it to provide most benefit for problems requiring good recall of facts, rather than reasoning abilities. %However, we see the opposite.
However, in Table~\ref{tab:summary_ratios} we see that retrieval is a better compute multiplier for STEM than for humanities or social sciences, and in Figure~\ref{fig:cat_impact} the gap between retrieval and base accuracy is also wider for STEM than for humanities.
Such knowledge may be harder to absorb during pre-training, and in contrast to long tail facts in SimpleQA, retrieval expanding the context may also function as additional processing rather than just storage. % speculation, probably delete?
% This may be because LLM reasoning uses high-level pattern matching, and retrieving relevant examples helps, or, it could be that being provided with related context conditions the model to be better at answering reasoning-based questions (i.e. it may be a weaker form of few-shot prompting).

% \begin{figure}
% \centering
% \begin{subfigure}{.5\textwidth}
%   \centering
%   \includegraphics[width=\linewidth]{category_subset_retrieval.pdf}
%   \caption{Categorical breakdown for subset retrieval}
%   \label{fig:sub1}
% \end{subfigure}%
% \begin{subfigure}{.5\textwidth}
%   \centering
%   \includegraphics[width=\linewidth]{category_full_retrieval.pdf}
%   \caption{Categorical breakdown for full retrieval}
%   \label{fig:sub2}
% \end{subfigure}
% \caption{MMLU Breakdown by category of impact of retrieval addition and compute budget}
% \label{fig:cat_impact}
% \end{figure}

To investigate this surprising observation, we calculated the increase in MMLU accuracy provided by full retrieval, for different subject areas, with a 6.4B model.
The top ten are shown in Table~\ref{tab:top_ten_retrieval_categories}. We also include the corresponding accuracy increase provided by subset retrieval.
Here we can see a mix of different subject types - those that might be expected to require good knowledge recall, such as law and medicine, along with those that might require logical reasoning, such as physics, and those that might require more abstract reasoning, such as philosophy.
A similar mixture of types of subjects may be seen for other model sizes, suggesting no strong correlation between subject type, model size, and the benefits of retrieval.

\begin{table*}[ht]
\centering
\caption{Top ten categories ordered by change in MMLU accuracy after introduction of full retrieval}
\label{tab:top_ten_retrieval_categories}
\small
\begin{tabular}{lcc}
\toprule
Subjects & Impact of full retrieval & Impact of subset retrieval \\
\midrule
Medical genetics & \textit{+21.1} & \textit{+19.5} \\
Miscellaneous & \textit{+19.2} & \textit{+17.8} \\
World religions & \textit{+18.5} & \textit{+19.6} \\
Philosophy & \textit{+17.9} & \textit{+17.1} \\
US foreign policy & \textit{+17.7} & \textit{+16.1} \\
International law & \textit{+17.0} & \textit{+17.3} \\
High school physics & \textit{+16.9} & \textit{+ \ 5.7} \\
Virology & \textit{+16.9} & \textit{+15.2} \\
College physics & \textit{+16.8} & \textit{+14.2} \\
College medicine & \textit{+16.6} & \textit{+11.8} \\
\bottomrule
\end{tabular}
\end{table*}

We also compared the MMLU answers between the 6.4B and 12.6B models, with and without retrieval, focusing on problems where the smaller base model gave an incorrect answer.
We identified problems where the addition of retrieval corrected the answer, but the increase in model size did not, and vice versa.
The main subjects of the former group were professional law, professional psychology, high-school macroeconomics, philosophy, and high-school mathematics.
The main subjects of the latter group were professional law, professional psychology, moral scenarios, elementary mathematics, and high school statistics.
Both groups have a mix of recall and reasoning problems, with much overlap. This suggests that there is not a strong bias for what kinds of problem retrieval can help with compared to increasing model size.

\iffalse
Although we did not find evidence that retrieval helps especially well in particular subject areas, we did find some differences in the benefits retrieval provides, versus scaling up pre-training.
Comparing the MMLU answers for the 12.6 billion parameter model to those of the 6.4 billion parameter model we observed:
\begin{itemize}
%    \item For $39.06\%$ of the problems increasing model size kept an answer correct
    \item For $8.69\%$ of the problems increasing model size introduced an error
    \item For $19.14\%$ of the problems increasing model size changed an incorrect answer to a correct one
    \item For $11.85\%$ of the problems increasing model size changed an incorrect answer to a different incorrect one
%    \item For $21.26\%$ of the problems increasing model size kept an incorrect answer
\end{itemize}

Comparing the MMLU answers for the 6.4 billion parameter model with and without retrieval, we observed:
\begin{itemize}
%    \item For $42.29\%$ of the problems adding retrieval kept an answer correct
    \item For $5.46\%$ of the problems adding retrieval introduced an error
    \item For $15.01\%$ of the problems adding retrieval changed an incorrect answer to a correct one
    \item For $7.61\%$ of the problems adding retrieval changed an incorrect answer to a different incorrect one
%    \item For $29.62\%$ of the problems adding retrieval kept an incorrect answer
\end{itemize}
\fi

In total, increasing model size changed answers for $39.7\%$ of MMLU problems, whereas adding retrieval changed answers for $28.1\%$ of the problems. A similar analysis for MATH-500 showed that increasing model size changed answers for $39.7\%$ of the problems, whereas adding retrieval changed answers for $28.7\%$ of the problems.
This suggests retrieval overall has less of an effect on model behavior than increasing model size.
Furthermore, it suggests that in areas where retrieval does not help, the problem is more that the model ignores the additional context, rather than the additional context being misleading.
Future work could involve encouraging models to better utilize retrieved context, possibly through prompt engineering or attention weighting.

\section{Test-Time Compute Experiments}

\begin{table*}[ht]
\centering
\caption{Comparing baseline reader model performance against retrieval with a variety of test-time compute options. All evaluations use chain-of-thought reasoning. We use Llama 3.1 8B instruct as the reader model. MMLU results are reported as macro average over subjects. VR refers to using variance reduction techniques such as MMR and bagging.}
\label{tab:main}
\begin{tabular}{lccccc}
\toprule
Method &  \makecell{MMLU \\ STEM} & \makecell{MMLU \\ Humanities} & \makecell{MMLU \\ Social} & \makecell{MMLU \\ Other} & \makecell{MMLU \\ All} \\
\midrule
Baseline            & 67.3 & 71.5 & 76.6 & 73.0 & 71.6 \\
w/ self-consistency & 72.3 & 74.8 & 79.3 & 76.4 & 75.3 \\
w/ retrieval        & 73.6 & 74.6 & 81.8 & 77.6 & 76.6 \\
w/ reranker         & 73.7 & 76.3 & 83.4 & 79.2 & 77.7 \\
\makecell[l]{w/ reranker + self-cons.} & 78.7 & 78.7 & 85.8 & 81.9 & 81.0 \\
\makecell[l]{w/ reranker + self-cons. + VR} & 80.2 & 79.5 & 87.4 & 82.3 & 82.1 \\
\bottomrule
\end{tabular}
\end{table*}

\begin{table*}[ht]
\centering
\caption{Continuation of Table~\ref{tab:main}. GPQA, and Math-500 results are over 10 trials.}
\label{tab:main2}
\begin{tabular}{lcccccc}
\toprule
Method &  SimpleQA & Math-500 & \makecell{GPQA \\ Bio.} & \makecell{GPQA \\ Chem.} & \makecell{GPQA \\ Phys.} & \makecell{GPQA \\ All} \\
\midrule
Baseline            & 1.5 & 48.7 & 46.2 & 26.4 & 28.3 & 30.6 \\
w/ self-consistency & N/A & 55.9 & 46.3 & 28.1 & 28.4 & 31.4 \\
w/ retrieval        & 65.7 & 56.7 & 45.1 & 27.3 & 34.0 & 33.2 \\
w/ reranker         & 74.0 & 56.8 & 46.7 & 28.6 & 36.0 & 34.8 \\
\makecell[l]{w/ reranker + self-cons.} & N/A & 64.3 & 48.5 & 30.1 & 36.8 & 36.1 \\
\makecell[l]{w/ reranker + self-cons. + VR} & N/A & 64.4 & 49.7 & 29.6 & 38.3 & 36.8 \\
\bottomrule
\end{tabular}
\end{table*}
% \makecell[l]{w/ reranker + self-cons. + VR} & N/A & 64.4 / 59.5 & 49.7 & 29.6 & 38.3 & 36.8 \\ 
% for math, 64.4 is without mmr, 59.5 is with mmr (probably just pick better number, mention in caption no MMR

% for simpleqa consistency, use N/A or repeat the standard number?
% how to handle decontamination for math? appendix?
% decide for math which checker? for now llama, put mini and decon in appendix
% decide for gpqa: diamond or main, olmes or standard
% for last option should we just do "VR", or breakdown MMR/LOO/Bagging. for now, just do bagging + MMR. but MMR hurts a lot for math

Knowing the limitations of learning with just pre-training, we attempt to better quantify the knowledge contained in these datasets by applying additional test-time compute on top of retrieval. If the model is able to answer a question with retrieval and test-time compute, the knowledge required to do so is likely in the dataset.
% but self-consistency alone can see gains - might be from previous pre-training data
% Need more justification on using Llama 3.1 8B instruct here. Why is it okay besides just practicality?
In this section, we use Llama 3.1 8B instruct \citep{llama} as the reader model due to its performance relative to its size, making it more practical to apply test-time compute. We then augment it with retrieval as described in Section~\ref{sec:setup}, and parallel inference with majority voting to select an answer (self-consistency). In addition to evaluating on MMLU, SimpleQA, and Math-500, we also evaluate on GPQA \citep{rein2024gpqa}.

Table~\ref{tab:main} and Table~\ref{tab:main2} show that the effects of self-consistency and retrieval are additive across all tasks, with the exception of SimpleQA where self-consistency does not help because it is purely a factuality benchmark.
Perhaps surprisingly, the two techniques help generally across all other tasks and sub-tasks, with little hint of specialization. Additionally, reranking seems to give a consistent boost on top of retrieval across tasks. Lastly, we take advantage of retrieving multiple documents and parallelizing trials by using older techniques like MMR \citep{carbonell1998use} to increase diversity, and bagging \citep{breiman1996bagging} (randomizing over a subset of documents) to reduce variance; these techniques give a further performance boost for MMLU and GPQA. 
% i use the word additive here, but can we argue for synergistic?
% do we need to explain why MMR seems to hurt for Math?
% mention these gains as somethign that may be learnable through better pre-training in future work?

If we view retrieval as a tool for the LLM, then our methods use test-time compute to improve the tool itself. This contrasts with self-consistency by itself, which parallelizes the model without additional enhancement, as well as with deep research, which in addition to parallelizing also uses test-time compute to use the tool for longer rather than to upgrade it. Retrieval is our vehicle through which we can put in additional compute, and do it in a data driven way.

\subsection{Learning from test-time compute vs pre-training}

\iffalse
\begin{table*}[ht]
\centering
\caption{Compute efficiency gains for each method relative to baseline performance. Values represent how many times more compute the baseline model would need to achieve the same performance as each method. Calculated using fitted power law equations for each MMLU category.}
\label{tab:efficiency}
\begin{tabular}{lccccc}
\toprule
Method & \makecell{MMLU \\ STEM} & \makecell{MMLU \\ Humanities} & \makecell{MMLU \\ Social} & \makecell{MMLU \\ Other} & \makecell{MMLU \\ All} \\
\midrule
Baseline            & 1.00× & 1.00× & 1.00× & 1.00× & 1.00× \\
w/ self-consistency & 2.45× & 2.81× & 2.12× & 2.11× & 2.14× \\
w/ retrieval        & 2.85× & 2.70× & 4.54× & 2.66× & 2.80× \\
w/ reranker         & 2.88× & 4.42× & 7.72× & 3.71× & 3.03× \\
\makecell[l]{w/ reranker + self-cons.} & 7.08× & 8.64× & 18.2× & 6.52× & 7.89× \\
\makecell[l]{w/ reranker + self-cons. + VR} & 9.05× & 10.5× & 34.5× & 7.03× & 10.2× \\
\bottomrule
\end{tabular}
\end{table*}
\fi
\begin{table*}[ht]
\centering
\caption{Compute efficiency gains for each method relative to baseline performance. Values represent how many times more compute the baseline model would need to achieve the same performance as each method. Calculated using fitted sigmoid equations for each MMLU category.}
\label{tab:efficiency}
\begin{tabular}{lccccc}
\toprule
Method & \makecell{MMLU \\ STEM} & \makecell{MMLU \\ Humanities} & \makecell{MMLU \\ Social} & \makecell{MMLU \\ Other} & \makecell{MMLU \\ All} \\
\midrule
Baseline            & 1.00× & 1.00× & 1.00× & 1.00× & 1.00× \\
w/ self-consistency & 2.62× & 1.93× & 1.63× & 2.21× & 2.10× \\
w/ retrieval        & 3.42× & 1.85× & 2.70× & 3.02× & 2.78× \\
w/ reranker         & 3.49× & 2.67× & 3.87× & 4.72× & 3.56× \\
\makecell[l]{w/ reranker + self-cons.} & 10.74× & 4.65× & 7.18× & 11.34× & 8.14× \\
\makecell[l]{w/ reranker + self-cons. + VR} & 15.72× & 5.68× & 11.66× & 13.15× & 11.10× \\
\bottomrule
\end{tabular}
\end{table*}

We can take the MMLU sigmoid fit from Section~\ref{sec:pretrain} to analyze Table~\ref{tab:main}. Although Llama 3.1 8B is trained at a much higher tokens per parameter ratio than the models in Section~\ref{sec:pretrain}, the sigmoid fit could still be reasonable because calculating the compute ratio between retrieval (with reranker) and base performance is within reason for MMLU (All). Additionally, estimates using our previous sigmoid fits would be a lower bound for compute multipliers because of diminishing returns at higher tokens per parameter counts. 

Despite the different pre-training dataset and Llama being significantly overtrained, we see that retrieval (with reranker) still functions as a 3.56x compute multiplier, similar to what would be expected of a model with the same MMLU base accuracy in our pre-training setup in Section~\ref{sec:pretrain}. Though we do not know the details of the Llama 3.1 pre-training dataset, it is likely that there is substantial overlap with the data we are retrieving from. Table~\ref{tab:efficiency} shows that, altogether, our methods provide at least an 11x compute multiplier over the pre-trained baseline.

We also see that the different test-time methods learn or utilize data differently from pre-training, as the multipliers are different across the categories. Even the different test-time methods have different behaviors, as both self-consistency and retrieval favor STEM and other, while the lift of reranker over retrieval favors humanities, social sciences, and other.
% also mention that llama has different pre-training dataset, its effect here

\subsection{A connection between retrieval and consistency}
% consistency can be used to show document quality / used as a reranker
While the previous results in this section demonstrate that self-consistency is a powerful tool for improving performance, it can also be used as an analytical tool for retrieval.
As displayed in Appendix~\ref{app:inter}, we can apply self-consistency on each individual document and rerank the documents with it. In Table~\ref{tab:inter} we see that inter-document consistency selects better top-1 documents than the reranker. However, this technique also requires a multiplicative number of additional trials, as previously we ran a fixed number of trials on all documents combined, but now we are doing it per document. We leave to future work ways to distill self-consistency into a more efficient reranker.

\begin{table*}[ht]
\centering
\caption{Inter-document consistency can act as a stronger reranker (k=1) than standard rerankers from \citet{qwen3embedding}; however, it requires calling the reader model many more times and is not compute efficient when compared to self-consistency on all documents at once.}
\label{tab:inter}
\begin{tabular}{lccccc}
\toprule
Reranker &  \makecell{MMLU \\ STEM} & \makecell{MMLU \\ Humanities} & \makecell{MMLU \\ Social} & \makecell{MMLU \\ Other} & \makecell{MMLU \\ All} \\
\midrule
Qwen3 Reranker 0.6B   & 71.2 & 72.9 & 77.7 & 74.1 & 73.7 \\
Inter-doc consistency & 75.4 & 75.4 & 82.7 & 78.1 & 77.6 \\
\bottomrule
\end{tabular}
\end{table*}

\section{Additional Analysis}
\subsection{Better pre-training datasets are not necessarily better retrieval datasets}

\begin{table*}[ht]
\centering
\caption{While FineWeb-edu is worse than DCLM when measuring pre-training performance, it is just as good if not slightly better for retrieval. Retrieval is on top of Llama 3.1 8B instruct with k=10. Pre-training numbers are 8B models trained for 1T tokens, as reported by \citet{su2024nemotron}.}
\label{tab:pretrain_vs_retrieval}
\begin{tabular}{lccc}
\toprule
Dataset &  \makecell{Pre-training \\ MMLU} & \makecell{Retrieval \\ MMLU} & \makecell{ Retrieval w/ reranker \\ MMLU}\\
\midrule
DCLM         & 53.4 & 74.5 & 76.4 \\
FineWeb-edu  & 42.9 & 75.2 & 76.6 \\\bottomrule
\end{tabular}
\end{table*}

Our retrieval datasets were built for the purpose of pre-training, which raises the question of whether better pre-training datasets make for better retrieval datasets. Table~\ref{tab:pretrain_vs_retrieval} suggests that this is not necessarily the case, as FineWeb-edu is worse than DCLM for pre-training, but is as good if not slightly better for retrieval. While DCLM contains more tokens than FineWeb-edu, prior work \citep{muennighoff2023,fang2025datasets} would suggest that this is not the reason for the gap in pre-training performance, while the size advantage should not be harmful for retrieval. We leave to future work how to determine the qualities that make a dataset good for pre-training or retrieval specifically.

\subsection{Importance of extraction and crawling} % (is it crawling though? could be filtering)}
\begin{table*}[ht]
\centering
\caption{Text extraction and crawling are important for creating good datasets. We vary the extraction done on top of Wikipedia, as well as how we expand the datastore. Retrieval uses reranker and is on top of Llama 3.1 8B instruct with k=6.}
\label{tab:extraction}
\begin{tabular}{lc}
\toprule
Dataset & SimpleQA\\
\midrule
Wikimedia Nov. 2023  & 55.4 \\
OLM June 2025  & 59.1 \\
Custom June 2025 & 69.0 \\
Custom + All Sources & 73.7 \\
Custom + Golden Links & 85.2 \\\bottomrule
\end{tabular}
\end{table*}
We investigate the importance of earlier stages of dataset creation through a case study of retrieval for SimpleQA. As constructed, over 70\% of the answers in SimpleQA can be found on Wikipedia. However, many works use out-of-date or pre-extracted versions of Wikipedia. This can lead to missing data, especially when the crucial piece of information comes from specialized elements.

We compare Wikimedia (Nov. 2023) \citep{wikidump} and OLM (June 2025) \citep{thrush2022pipeline}, two of the most popular Wikipedia extractions on HuggingFace, against a custom extracted Wikipedia (June 2025) described in Appendix~\ref{app:extraction}.
Qualitatively, we find that existing extractions often fail to extract elements like bullet points, tables, and info boxes. Table~\ref{tab:extraction} quantitatively demonstrates this with up to a 13.6 percentage point difference in SimpleQA performance by simply changing the version of the Wikipedia dataset.

Next, we compare expanding the retrieval datastore by adding our other sources against adding non-Wikipedia golden links provided by SimpleQA. Table~\ref{tab:extraction} shows that there is an 11.5 percentage point difference, and we find that only a small fraction of the non-Wikipedia golden links are present in CommonCrawl. This suggests that open-source datasets could be further improved at the web crawling stage.
% and we find that less than one percent of the non-Wikipedia golden links are present in CommonCrawl. This suggests that open-source datasets could be further improved at the web crawling stage.
% TODO is it really crawling? may need to change from crawling to filtering because it got filtered out? 

\subsection{Robustness when scaling retrieval data}
% Is this interesting?
\begin{figure*}[htbp]
    \centering
    \includegraphics[width=1.0\linewidth]{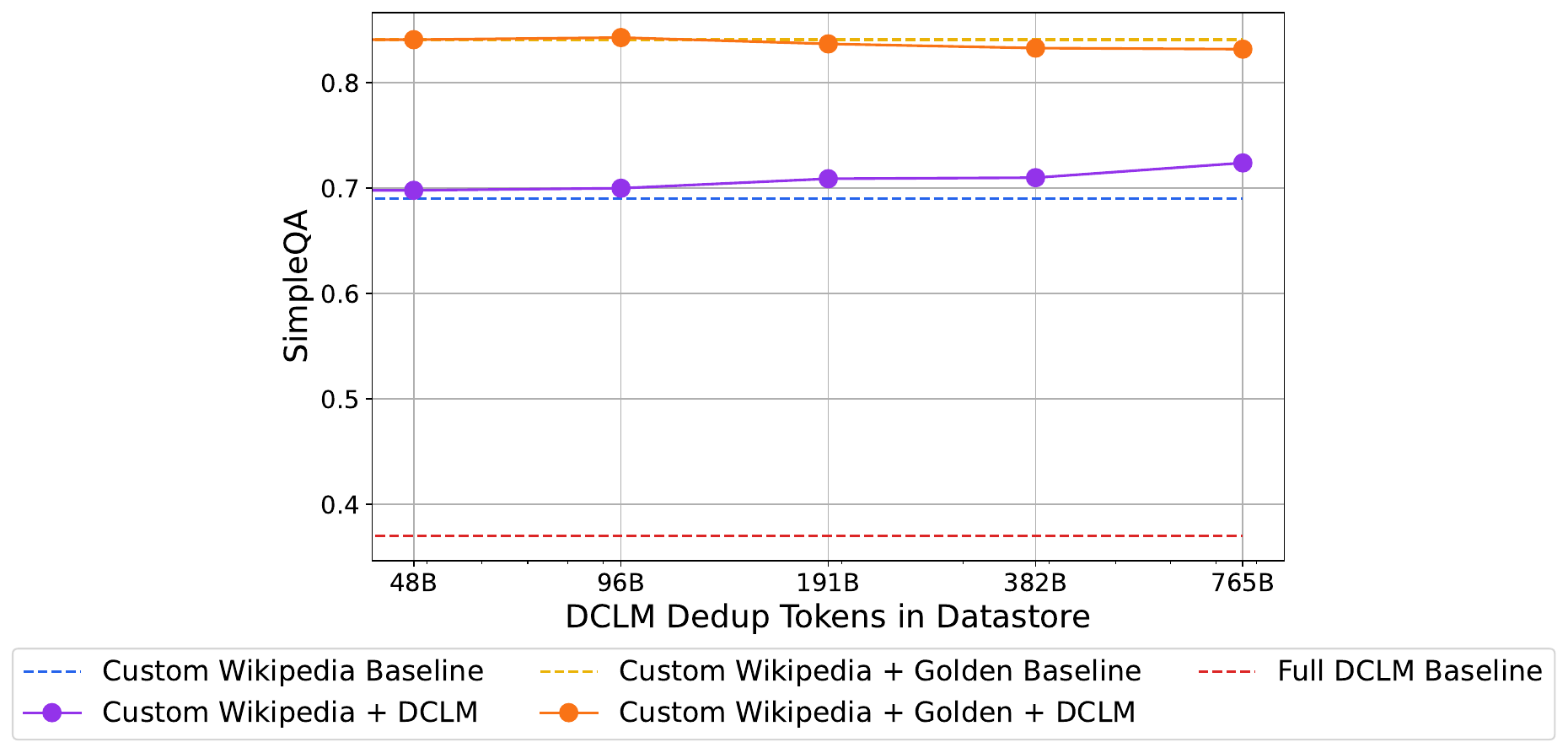}
    \caption{For SimpleQA, our retrieval system is fairly robust to scaling the retrieval datastore, even if the new data does not contain useful information. Our custom Wikipedia contains 22B tokens, and additional DCLM data helps a little, or when also starting with additional golden link data, hurts only a little.}
    \label{fig:wiki_scale}
\end{figure*}
% Maybe DCLM line should be scaling instead of baseline for all

Given that most of SimpleQA can be solved with Wikipedia and the additional provided golden links, we investigate the effect of additional web crawl data on the retrieval system. In Figure~\ref{fig:wiki_scale} we see that the additional data has only a small distracting effect, as the accuracy on SimpleQA stays close to the baseline. However, we acknowledge that SimpleQA measures factual knowledge, and the scaling effect may be different for reasoning tasks.

% \subsection{SimpleQA count checker?}
% requires some additional work. I have a dictionary where for the top 25 documents for each SimpleQA question I check if k=1 gets it right. We then compare this to which questions are solved through pre-training

\section{Future work}
We have shown that pre-training does not fully utilize all the knowledge contained within today's open-source pre-training datasets. This would suggest that there are still many algorithmic improvements left to explore. Additionally, in our process of analyzing data quality, we have also shown that there is room for improving datasets, at the very least in terms of crawling and extraction.

Within this work, we explore a limited number of simple test-time techniques on a limited set of evaluations. However, it is quite likely that applying advanced techniques like query rewriting, test-time training, and reinforcement learning for retrieval will further boost the performance on the same datasets \citep{hardt2023test,ma2023query}. We also believe that these findings apply to even broader domains. Initial results in Appendix~\ref{app:code} suggest that retrieving from pre-training datasets also benefits code generation.

Beyond improvements, we would also like to better understand how data is used during pre-training. Our measurement of retrieval as a compute multiplier parallels that of well-tuned Mixture-of-Experts models \citep{clark2022unified}. Additional exploration in this area could uncover whether these two methods have significant overlap in terms of data usage.

% Overall conclusion, answer the question in the intro
% Improving Pre-training datasets
% Improving Pre-training (architectures?) to get more immediate performance from datasets
% More complicated Test-Time Compute
% Relationship with MoE
% More tasks? Iniital LiveCodeBench numbers?
% Probably not, but can consider discussing templating

% Acknowledgements, running list: Sewon, Rulin?, Aniket, Apple Readers?

% \subsubsection*{Acknowledgments}
% Use unnumbered third level headings for the acknowledgments. All
% acknowledgments, including those to funding agencies, go at the end of the paper.
\subsection*{Acknowledgments}
We would like to thank Sewon Min, Rulin Shao, and Anikait Singh for helpful conversations and feedback at various stages of the project.

\bibliography{iclr2026_conference}
\bibliographystyle{iclr2026_conference}

%%%%%%%%%%%%%%%%%%%%%%%%%%%%%%%%%%%%%%%%%%%%%%%%%%%%%%%%%%%%
\newpage

\appendix

% Nice to have: gpqa format, gpqa diamond, gpqa google / rewrite

\section{Pre-training details}
\label{app:pretraining}
% We follow a recipe very similar to that open-sourced as the \hyperref[https://github.com/apple/axlearn/blob/main/axlearn/experiments/text/gpt/honeycrisp.py]{"Honeycrisp"} model series in the Axlearn training framework \cite{lee2025axlearnmodularlargemodel}. % remove link for submission
We follow a recipe very similar to that open-sourced as the "Honeycrisp" model series in the Axlearn training framework \cite{lee2025axlearnmodularlargemodel}.

We train each dense decoder model for around 20 tokens per parameter, and follow a cosine-with-linear-warmup learning rate schedule with a peak learning rate of 1e-2, decaying to 0.01 of the peak lr. As described in the "Honeycrisp" model definitions, we use a muP-style parameterization to achieve learning rate transfer as we scale up the models. 
The model architecture is similar to the LLaMA series, with Swi-GLU FFNs, RoPE positional encodings, and Grouped Query Attention (GQA) using a key/value-to-query ratio of 1:8.

\begin{table}[h]
\centering
\begin{tabular}{ccc}
\toprule
Compute Budget & Paramters & Tokens \\
\midrule
$5.64 \times 10^{21}$ & 6.4B & 147B \\
$1.90 \times 10^{22}$ & 12.6B & 252B  \\
$7.04 \times 10^{22}$ & 23.3B & 503B \\
$1.74 \times 10^{23}$ & 36.8B & 786B \\
$7.34 \times 10^{23}$ & 77.8B & 1573B \\
\bottomrule
\end{tabular}
\label{tab:multiplier_sigmoid}
\end{table}

\iffalse
\section{Compute multiplier with sigmoid function}
\label{app:sigmoid}
\begin{table}[h]
\centering
\caption{We fit the base model performance to a sigmoid function with bounds to get $y = 0.25 + \dfrac{0.6907}{1 + \exp\!\big(-0.7968 \cdot (\log_{10}(x) - \log_{10}(2.48 \times 10^{22}))\big)}$, where 0.25 is the random baseline and 0.9407 is the maximum achievable accuracy \citep{gema2024we}. We use this equation to measure retrieval as a compute multiplier for the base model. The average compute ratio is 4.86x, the geometric mean is 4.66x, and the median is 4.74x.} % maybe 0.9351 is micro average, check what macro should be, maybe 0.941
% macro average should be 0.9407; by sub is STEM: 95.44, humanities 93.77, social sciences 95.75, other 91.14
\begin{tabular}{ccccc}
\toprule
Compute Budget & \makecell{Baseline \\ MMLU} & \makecell{Retrieval \\ MMLU} & \makecell{Compute for base \\ to match retrieval} & Compute Ratio \\
\midrule
$5.64 \times 10^{21}$ & 0.4873 & 0.6063 & $2.98 \times 10^{22}$ & 5.28x \\
$1.90 \times 10^{22}$ & 0.6021 & 0.6943 & $1.36 \times 10^{23}$ & 7.17x \\
$7.04 \times 10^{22}$ & 0.6623 & 0.7410 & $3.34 \times 10^{23}$ & 4.74x \\
$1.74 \times 10^{23}$ & 0.7107 & 0.7775 & $7.35 \times 10^{23}$ & 4.23x \\
$7.34 \times 10^{23}$ & 0.7633 & 0.8186 & $2.11 \times 10^{24}$ & 2.88x \\
\bottomrule
\end{tabular}
\label{tab:multiplier_sigmoid}
\end{table}
\fi

\section{Decontamination}
\label{app:decontamination}
We perform n-gram decontamination against our test sets using n-grams in token-space (according to our tokenizer).
We drop entire documents on collision with a single 16-gram from the MMLU test-set or 26-gram for Math-500. Through visual examination we found that shorter n-gram overlaps were too aggressive (noting that our tokenizer e.g. splits numbers into single digits).

\section{Custom extraction}
\label{app:extraction}
We implement a custom HTML extraction pipeline and apply it to all pages from the Wikipedia domain found in our general web-crawl.
Specifically, we first apply a lightweight pre-processing step to remove script, style, unmatched meta tags, HTML comments, links, and images. We then use the ReaderLM-v2 \citep{wang2025readerlmv2smalllanguagemodel} to extract the coarsely simplified HTML into structured plain-text.

We note that this approach improves on recall (especially for tables and some information-boxes) over publicly available Wikipedia extractions, including the one recently provided by the Wikimedia organization \cite{structuredwikimedia}.

\section{Detokenize subset vs random subset}
\label{app:detokenize_subset}
\begin{table}[h]
\centering
\caption{In Figure~\ref{fig:figure1} we show the effect of retrieving from a subset similar to that seen in pre-training for that compute budget. Specifically, we take a random subset that is the same size of what is seen during pre-training for each source. In this table, we compare at the smallest $5.64 \times 10^{21}$ compute budget between random subset and detokenizing the exact pre-training data and using that for retrieval. The results are fairly similar, and the larger gap in Math-500 may be attributed to randomness or contamination. Note that this gap will shrink as the compute budget increases because the two datasets will have increasing overlap.}
\label{tab:detokenize_subset}
\begin{tabular}{lccccc}
\toprule
Data & MMLU & Math-500 & SimpleQA\\
\midrule
Random subset & 59.0 & 32.0 & 35.5 \\
Exact subset  & 58.6 & 26.8 & 33.9 \\
\bottomrule
\end{tabular}
\end{table}
\newpage

\section{Fits by MMLU Categories}

\begin{table}[h]
\centering
\caption{Pre-training vs. retrieval compute ratios for STEM. Sigmoid fit (min=0.25, max=0.9544):
$y = 0.25 + \dfrac{0.7044}{1 + \exp\!\big(-0.7351 \cdot (\log_{10}(x) - 22.9965)\big)}$.
Average compute ratio is 6.16×, geometric mean is 5.78×.}
\label{tab:stem_ratios}
\begin{tabular}{ccccc}
\toprule
Compute Budget & \makecell{Baseline \\ MMLU} & \makecell{Retrieval \\ MMLU} & \makecell{Compute for base \\ to match retrieval} & Compute Ratio \\
\midrule
$5.64 \times 10^{21}$ & 0.4247 & 0.5493 & $3.85 \times 10^{22}$ & 6.82× \\
$1.90 \times 10^{22}$ & 0.5367 & 0.6399 & $1.94 \times 10^{23}$ & 10.23× \\
$7.04 \times 10^{22}$ & 0.5788 & 0.6749 & $3.68 \times 10^{23}$ & 5.23× \\
$1.74 \times 10^{23}$ & 0.6376 & 0.7197 & $8.71 \times 10^{23}$ & 5.01× \\
$7.34 \times 10^{23}$ & 0.7056 & 0.7705 & $2.58 \times 10^{24}$ & 3.52× \\
\bottomrule
\end{tabular}
\end{table}

\begin{table}[h]
\centering
\caption{Pre-training vs. retrieval compute ratios for Humanities. Sigmoid fit (min=0.25, max=0.9377):
$y = 0.25 + \dfrac{0.6877}{1 + \exp\!\big(-0.8008 \cdot (\log_{10}(x) - 22.1259)\big)}$.
Average compute ratio is 2.52×, geometric mean is 2.44×.}
\label{tab:humanities_ratios}
\begin{tabular}{ccccc}
\toprule
Compute Budget & \makecell{Baseline \\ MMLU} & \makecell{Retrieval \\ MMLU} & \makecell{Compute for base \\ to match retrieval} & Compute Ratio \\
\midrule
$5.64 \times 10^{21}$ & 0.5277 & 0.6013 & $1.51 \times 10^{22}$ & 2.69× \\
$1.90 \times 10^{22}$ & 0.6205 & 0.6847 & $6.33 \times 10^{22}$ & 3.33× \\
$7.04 \times 10^{22}$ & 0.6961 & 0.7398 & $1.81 \times 10^{23}$ & 2.57× \\
$1.74 \times 10^{23}$ & 0.7420 & 0.7788 & $4.24 \times 10^{23}$ & 2.44× \\
$7.34 \times 10^{23}$ & 0.7871 & 0.8169 & $1.14 \times 10^{24}$ & 1.55× \\
\bottomrule
\end{tabular}
\end{table}

\begin{table}[h]
\centering
\caption{Pre-training vs. retrieval compute ratios for Social Sciences. Sigmoid fit (min=0.25, max=0.9575):
$y = 0.25 + \dfrac{0.7075}{1 + \exp\!\big(-0.9563 \cdot (\log_{10}(x) - 21.9772)\big)}$.
Average compute ratio is 3.51×, geometric mean is 3.42×.}
\label{tab:social_ratios}
\begin{tabular}{ccccc}
\toprule
Compute Budget & \makecell{Baseline \\ MMLU} & \makecell{Retrieval \\ MMLU} & \makecell{Compute for base \\ to match retrieval} & Compute Ratio \\
\midrule
$5.64 \times 10^{21}$ & 0.5419 & 0.6555 & $1.93 \times 10^{22}$ & 3.42× \\
$1.90 \times 10^{22}$ & 0.6770 & 0.7538 & $8.39 \times 10^{22}$ & 4.42× \\
$7.04 \times 10^{22}$ & 0.7538 & 0.8192 & $2.86 \times 10^{23}$ & 4.07× \\
$1.74 \times 10^{23}$ & 0.7983 & 0.8501 & $5.97 \times 10^{23}$ & 3.43× \\
$7.34 \times 10^{23}$ & 0.8375 & 0.8828 & $1.63 \times 10^{24}$ & 2.22× \\
\bottomrule
\end{tabular}
\end{table}

\begin{table}[h]
\centering
\caption{Pre-training vs. retrieval compute ratios for Other. Sigmoid fit (min=0.25, max=0.9114):
$y = 0.25 + \dfrac{0.6614}{1 + \exp\!\big(-0.8008 \cdot (\log_{10}(x) - 22.2759)\big)}$.
Average compute ratio is 9.27×, geometric mean is 8.96×.}
\label{tab:other_ratios}
\begin{tabular}{ccccc}
\toprule
Compute Budget & \makecell{Baseline \\ MMLU} & \makecell{Retrieval \\ MMLU} & \makecell{Compute for base \\ to match retrieval} & Compute Ratio \\
\midrule
$5.64 \times 10^{21}$ & 0.4834 & 0.6418 & $5.53 \times 10^{22}$ & 9.80× \\
$1.90 \times 10^{22}$ & 0.6046 & 0.7222 & $2.62 \times 10^{23}$ & 13.78× \\
$7.04 \times 10^{22}$ & 0.6601 & 0.7598 & $6.17 \times 10^{23}$ & 8.77× \\
$1.74 \times 10^{23}$ & 0.7006 & 0.7882 & $1.31 \times 10^{24}$ & 7.53× \\
$7.34 \times 10^{23}$ & 0.7516 & 0.8271 & $4.76 \times 10^{24}$ & 6.48× \\
\bottomrule
\end{tabular}
\end{table}
\newpage
\section{Math-500 checker for USC}
\label{app:math_checker}
\begin{table}[h]
\centering
\caption{Math-500 answers are open-ended so we use universal self-consistency (USC) \citep{chen2023universal} in Table~\ref{tab:main2} with the reader model itself as the checker (Llama 3.1 8B). Here, we compare this against using GPT-4.1 mini as the USC checker model.}
\label{tab:math_checker}
\begin{tabular}{lcc}
\toprule
Method & \makecell{Llama 3.1 8B \\ checker} & \makecell{GPT-4.1 mini \\ checker} \\
\midrule
Baseline & 48.7 & N/A \\
w/ self-consistency & 55.9 & 62.2 \\
w/ retrieval & 56.7 & N/A \\
w/ reranker & 56.8 & N/A \\
w/ reranker + self-cons. & 64.3 & 69.7 \\
w/ reranker + self-cons. + VR & 64.4 & 71.8 \\
\bottomrule
\end{tabular}
\end{table}

% \section{GPQA format}

% \section{GPQA Diamond}

% \section{GPQA Google}
% \label{app:google}

\section{LiveCodeBench Results}
\label{app:code}
\begin{table}[h!]
\centering
\caption{Retrieving from the python portions of the Stack v2 \citep{lozhkov2024starcoder} and CommitPack \citep{muennighoff2023octopack} to augment generation for LiveCodeBench Code Generation \citep{jain2024livecodebench}.}
\label{tab:code}
\begin{tabular}{lcc}
\toprule
Model & Baseline & Retrieval (k=3) \\
\midrule
gpt-4o-2024-08-06 & 0.3793 & 0.4276 \\
\bottomrule
\end{tabular}
\end{table}

\newpage
\section{Inter-document consistency}
\label{app:inter}
\begin{figure}[h!]
  \centering
  % Uncomment and replace with your figure
  \includegraphics[width=0.8\linewidth]{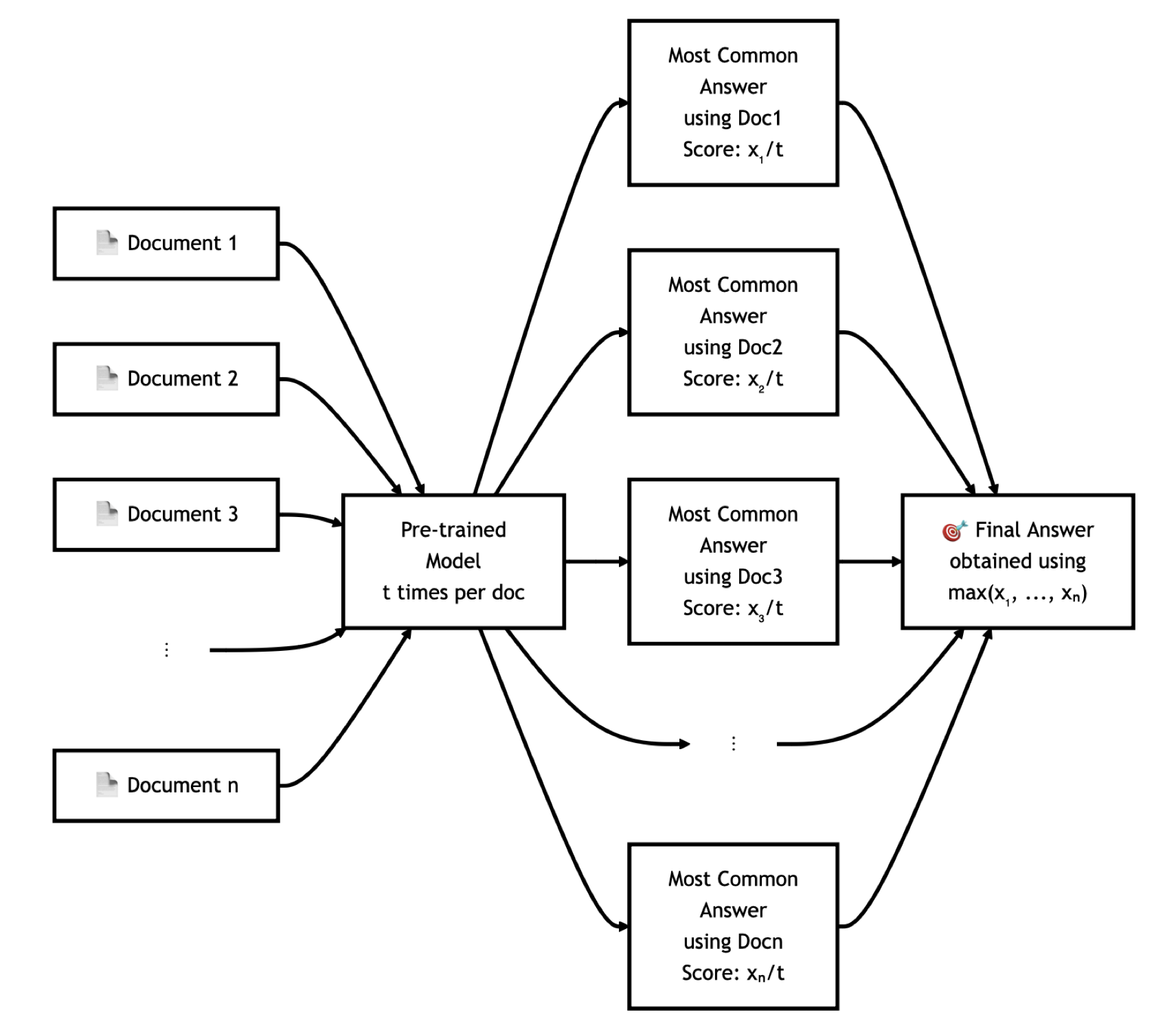}
  %\fbox{\rule{0pt}{4cm}\rule{14cm}{0pt}}
  \caption{Inter-document consistency can be used to analyze retrieval and consistency. We apply self-consistency on generating while retrieving from individual documents, and select the answer from the most self-consistent document.}
  \label{fig:consistency}
\end{figure}

\end{document}